\documentclass{article}

\usepackage[T1]{fontenc}
\usepackage{iclr2027_conference,times}
\usepackage[T1]{fontenc}

\usepackage{amsmath,amsfonts,bm}

\def\eqref#1{equation~\ref{#1}}

\def\1{\bm{1}}

\def\vtheta{{\bm{\theta}}}

\def\vp{{\bm{p}}}

\def\vx{{\bm{x}}}
\def\vy{{\bm{y}}}

\def\mI{{\bm{I}}}

\DeclareMathAlphabet{\mathsfit}{\encodingdefault}{\sfdefault}{m}{sl}
\SetMathAlphabet{\mathsfit}{bold}{\encodingdefault}{\sfdefault}{bx}{n}

\DeclareMathOperator*{\argmax}{arg\,max}
\DeclareMathOperator*{\argmin}{arg\,min}

\usepackage{xcolor}
\usepackage{hyperref}
\usepackage{url}

\usepackage{amsmath,amssymb,amsfonts,bm}
\let\argmin\relax
\DeclareMathOperator*{\argmin}{\mathsf{arg\,min}} %
\let\argmax\relax
\DeclareMathOperator*{\argmax}{\mathsf{arg\,max}}
\usepackage{booktabs}
\usepackage{pifont,makecell}

\usepackage{multirow}
\usepackage{array}
\usepackage{graphicx}
\usepackage{algorithm}
\usepackage{algpseudocode}
\usepackage{caption}
\usepackage{float}
\usepackage{enumitem}
\usepackage{adjustbox}
\usepackage{microtype}
\usepackage{mathtools}
\usepackage{etoolbox}
\usepackage{placeins}

\makeatletter
\newenvironment{ruledalgorithm}[2][]{%
  \par\addvspace{\medskipamount}\noindent
  \begin{minipage}{\linewidth}%
  \refstepcounter{algorithm}%
  \ifx\relax#1\relax\else\label{#1}\fi
  \hrule height 0.8pt\relax
  \vspace{0.45ex}
  {\raggedright\textbf{Algorithm \thealgorithm}~#2\par}
  \vspace{0.35ex}\hrule height 0.5pt\relax
  \vspace{0.4ex}
  \begin{algorithmic}[1]
}{%
  \end{algorithmic}
  \vspace{0.35ex}\hrule height 0.8pt\relax
  \end{minipage}%
  \par\addvspace{\medskipamount}%
}
\makeatother

\newcommand{\papertitle}{Mutually Adversarial Self-Training with Evolving Data for Unified Multimodal Models}

\iclrfinalcopy
\hypersetup{
  pdftitle={Mutually Adversarial Self-Training with Evolving Data for Unified Multimodal Models},
  pdfauthor={Wentao Zhou, Weijie Gan, Jiayun Wang},
  pdfsubject={Unified multimodal understanding and generation},
  pdfkeywords={unified multimodal models, self-training, self-play}
}

\title{\papertitle}

\author{
Wentao Zhou\textsuperscript{1,}\thanks{Equal contribution.}\quad
Weijie Gan\textsuperscript{2,}\footnotemark[1]\quad
Jiayun Wang\textsuperscript{1}\\[0.5ex]
{\normalfont\textsuperscript{1}Georgia Institute of Technology}\\
{\normalfont\textsuperscript{2}Washington University in St. Louis}
}

\begin{document}

\maketitle
\pagestyle{plain}
\thispagestyle{plain}

\begin{abstract}
Unified multimodal models (UMMs) combine image generation and visual understanding in a shared backbone. Since generation and understanding are inverse tasks, recent studies self-train UMMs by letting the two branches \emph{cooperatively} supervise each other. We introduce MATE (\textit{Mutually Adversarial self-Training with Evolving data}), a reinforcement-learning-based post-training framework in which the two branches instead \emph{challenge} each other, and the challenges evolve as the model trains. MATE lets generation and understanding take turns to be challenger and solver. Given an image, the understanding branch proposes several candidate descriptions that the generation branch must turn back into similar images, and vice versa. The candidates are screened for consistency with the image or prompt they were proposed from, and the solver is trained on the candidate it handles worst. The adversary thus comes from the model's own outputs, and no separate adversary is trained. Moreover, the candidates that defeat one branch become the sources of the next challenges to the other in the next epoch, which keeps the challenges evolving with the model and turns the training into self-play in data space. On Janus-Pro-1B, MATE improves GenEval by 2.4 points, DPG-Bench by 1.7 points, and the average over nine understanding benchmarks by 0.7 points, while strengthening consistency across repeated image-text cycles.
\end{abstract}

\section{Introduction}\label{sec:intro}
\emph{Unified multimodal models (UMMs)} integrate visual understanding and generation into a single backbone~\citep{zhao2025unified,xie2024unifying}, promising mutually reinforcing capabilities and a single interface for both. UMM training still follows the pretrain-then-finetune recipe, so UMM capabilities are bounded by the coverage of externally annotated data. Self-training~\citep{deng2025selfimprovementsurvey,yang2026selfimprovement}, on the other hand, is a natural way past this bound, and the unified architecture opens a new route to it. Understanding and generation are inverse tasks handled by two branches of one model, so each branch can supervise the other without new labels. Recent work builds on this duality by letting one branch score, filter, or supply training data for the other~\citep{hong2025suder,qu2024silmm,jin2025srum,yang2025hermesflow,gvu2026,mao2025unirl,han2026unicorn}. In most of such work, the two branches \emph{cooperate}, agreeing on what counts as good and helping each other do more of it on whatever examples happen to be sampled.

In this paper, we argue that the two branches can also \emph{challenge} each other. First, adversarial pressure directs learning toward cases on which the model performs poorly, whereas a cooperative loop tends to reinforce what the model already does well. Second, a UMM is the natural place for such a contest. Its two capabilities are inverses of each other, so each branch is a ready-made adversary for the other. Moreover, because they share one backbone, what one learns from its hard cases is, at least in principle, available to the other.

Our key idea is therefore to \emph{pit the two branches of a UMM against each other, with generation and understanding taking turns as challenger and solver}. The generator's images challenge the understanding branch to describe them, and the understanding branch's descriptions challenge the generator to realize them. Some adversarial frameworks introduce a separately trained component, such as a discriminator or perturber~\citep{saxena2021gans,su2025unigame}. Here, neither branch is trained to challenge. Each produces its ordinary outputs, and the challenge is whichever of them the other branch handles worst, so the adversary lives in \emph{data} and consists only of these selected outputs.

The next question is how an adversary made of data can supply a continuing training signal, since a selected challenge is consumed by one update and then spent. We take from self-play~\citep{zhao2025azr,huang2025rzero} the principle that the opponent improves as the player does. Unlike prior work, which keeps the opponent in a second network~\citep{goodfellow2014gan} or an archive of past checkpoints~\citep{vinyals2019alphastar,berner2019dota}, we propose to keep it in an evolving input pool of the model's own selected outputs. The image that defeated the understanding branch becomes the source of the next challenges to the generator, and vice versa. This evolution keeps the challenge moving with the model, whereas a fixed pool pins it near where it started and runs out of pressure once its sources are mastered. We call this \emph{self-play in data space}.

\begin{figure}[t]
\centering
\includegraphics[width=\linewidth]{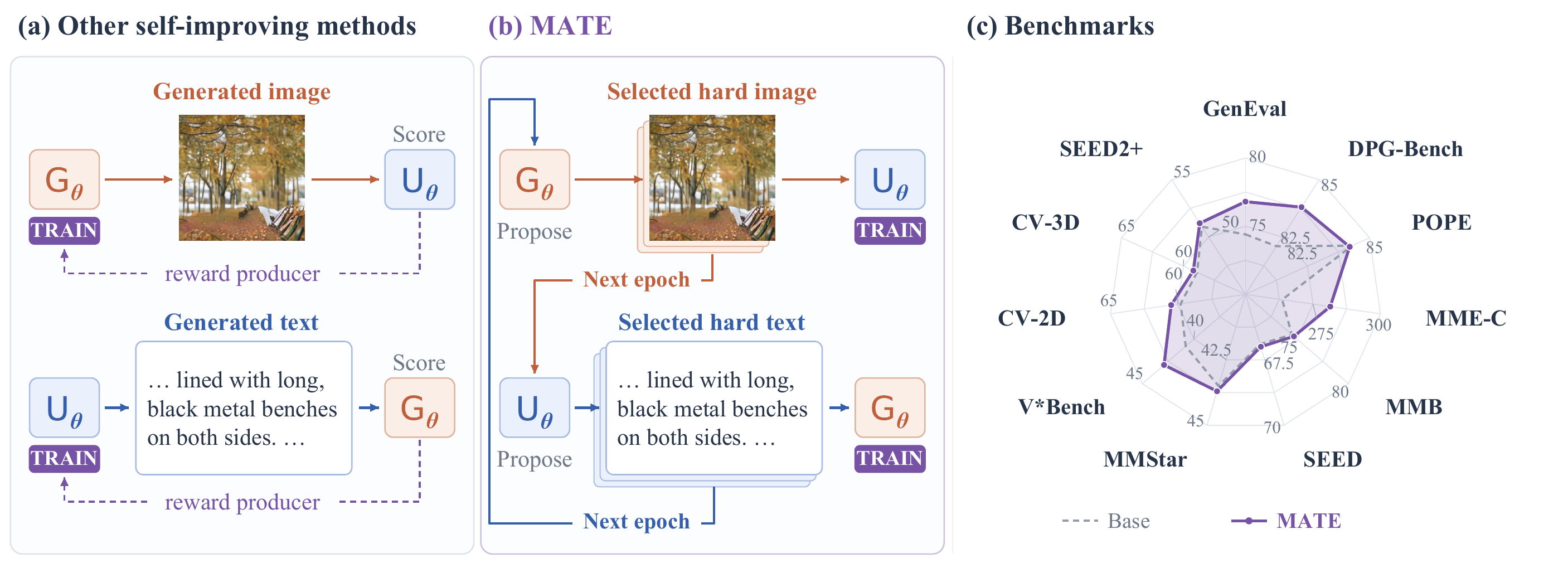}
\caption{\textbf{Self-improvement through mutual challenges.}
(a) In other self-improving methods such as SUDER~\citep{hong2025suder}, one
branch scores an output to reward its producer. (b) Our MATE selects valid outputs
that challenge the opposite branch, trains that solver, and uses the selected
outputs to construct new challenges in the next epoch. In both panels,
$\mathsf{G}_{\vtheta}$ and $\mathsf{U}_{\vtheta}$ share one backbone and the
examples are real training records.
(c) MATE and Janus-Pro-1B on the 11 benchmarks of the main tables. Each axis
uses its own truncated linear range, shared by both methods, so polygon areas
do not measure overall performance. MME-C is in cognition points, and other
scores are percentages.}
\vspace{-5mm}
\label{fig:abstract}
\end{figure}

{\bf Our approach:} We propose \emph{Mutually Adversarial self-Training with Evolving data} (MATE), a post-training framework that combines mutual challenge between the two branches of a UMM with self-play in data space (Figure~\ref{fig:abstract}). MATE keeps the objective and the optimizer of standard reinforcement-learning post-training and changes only where the inputs of each branch come from and what its outputs are scored against. Each epoch plays one round at every source in the input pools. A round starts from a source text or a source image, and the challenger proposes several candidates in the other modality. Only the candidates that agree with the source are admitted, so a candidate the solver fails on is hard rather than invalid. The solver attempts each admitted candidate several times, and each attempt is scored on its match to the candidate and its consistency with the source. The candidate with the lowest mean score is the challenge, and the solver is updated with group-relative policy optimization (GRPO)~\citep{shao2024deepseekmath} on the attempts that measured it. The selected challenges of one epoch are the sources of the next, so an output that defeated one branch becomes the source of new challenges to the other. Our contributions are:
\begin{itemize}
    \item \textbf{Adversarial self-training for UMMs.} To our knowledge, MATE is the first post-training framework in which the two branches of a UMM challenge each other, with both branches trained and no adversary beyond the model itself.
    \item \textbf{Self-play in data space.} We propose a self-play scheme that evolves the model's opponent in data, as a pool of its own hardest admitted outputs, rebuilt every epoch from the outputs that defeated one branch to challenge the other.
    \item \textbf{Empirical study.} On Janus-Pro-1B~\citep{chen2025januspro}, MATE improves generation and average understanding over the base model and outperforms two self-training baselines on both. Ablations indicate that challenge selection and evolving pools contribute to the generation gains.
\end{itemize}

\section{Related Work}\label{sec:related}
We relate MATE to four lines of prior work, and Table~\ref{tab:positioning} in Appendix~\ref{app:positioning} compares it with the closest methods along the axes discussed below.

\paragraph{Unified multimodal models.}
UMMs place visual understanding and image generation in one backbone~\citep{zhao2025unified,xie2024unifying,chen2024nexttoken,wang2025multimodalgenai}. They differ in how they represent and generate images. Fully autoregressive models predict discrete visual tokens in sequence~\citep{team2024chameleon,wang2024emu3,chen2025januspro}. Show-o~\citep{xie2024showo} combines autoregression with discrete diffusion over those tokens inside a single transformer. Other designs add a flow-matching head to the backbone~\citep{xie2025showo2}, connect a pretrained multimodal model to a diffusion model~\citep{wu2025openuni}, or assign understanding and generation to separate transformer experts~\citep{deng2025bagel}. MATE uses a backbone that both produces images and describes them; our experiments use the fully autoregressive Janus-Pro.

\paragraph{Cooperative self-training for UMMs.}
Self-training without new annotation is studied for language models~\citep{yang2026selfimprovement,tao2024selfevolution} and for multimodal models~\citep{deng2025selfimprovementsurvey}. In a UMM the generation and understanding branches are inverses, so each can supply supervision for the other. SUDER~\citep{hong2025suder} and GvU~\citep{gvu2026} score an output by the likelihood that the reverse task assigns to the original input. SUDER applies this in both directions and trains both branches, whereas GvU scores only generated images and trains only the generator. UniRL~\citep{mao2025unirl}, SILMM~\citep{qu2024silmm}, and HermesFlow~\citep{yang2025hermesflow} check a generated image against questions whose answers follow from the prompt. UniRL constructs the prompts and trains both branches on the correctness of the answers, SILMM lets the model write both the prompts and the questions, and HermesFlow trains the generator to prefer the image with the most correct answers over the image with the fewest. SRUM~\citep{jin2025srum} and UniCorn~\citep{han2026unicorn} have the understanding branch rate the generated images. SRUM rates an image as a whole and region by region, and UniCorn also lets the model write the prompts and fine-tunes on its best-rated images together with captioning, judgment, and reflection data. These methods use cross-modal agreement to score outputs or build preference pairs. MATE selects source-consistent candidates that the opposite branch handles poorly, then carries these challenges into the next epoch's source pools.

\paragraph{Adversarial training.}
Adversarial training exposes a model's weaknesses with challenges that are optimized against it~\citep{goodfellow2014gan,goodfellow2015adversarial,madry2018pgd,razaviyayn2020nonconvex} or selected from a pool because the model already finds them hard~\citep{shrivastava2016ohem}. On understanding tasks, recent multimodal work obtains such challenges from a challenger model that mines hard negatives for a solver~\citep{qiu2026duel} or from a multi-player game~\citep{wang2025visionzero}. For UMMs, UniGame~\citep{su2025unigame} attaches a perturber at the shared token interface, constrains its perturbations around the embeddings of real images, retains hard examples in a buffer, and trains the understanding branch on labeled question-answer pairs. MATE constructs challenges in both directions and optimizes both tasks, whereas UniGame optimizes the understanding objective and a learned perturber. MATE draws its challenges from candidates the branches already produce and needs no annotation beyond the source itself, so it introduces no component beyond the model.

\paragraph{Self-play and evolving training data.}
Self-play trains a model against opponents derived from itself~\citep{tesauro1995tdgammon,silver2017alphagozero,chen2024spin}, and methods differ in where the opponent is kept. It can be a population of current and past agents~\citep{vinyals2019alphastar,berner2019dota}, a proposer trained to pose problems for the solver~\citep{zhao2025azr,huang2025rzero}, or a buffer of training cases that is replayed and expanded~\citep{jiang2021plr,jiang2021robustplr,parkerholder2022accel}. EVA~\citep{ye2024eva} is the closest of the buffer methods to MATE, since it evolves its most informative prompts into new ones. Self-play is already claimed for UMM post-training~\citep{yang2025hermesflow,han2026unicorn,su2025unigame}, where the opponent is a role, a module, or an earlier round of the same model. MATE instead keeps the opponent in its input pools, where a description or an image that one branch handles worst becomes a source of challenges for the other. EVA returns its evolved prompts to the same solver in the same modality, whereas in MATE both alternate from one epoch to the next.

\section{Method}\label{sec:method}
MATE trains the generation and understanding branches of a UMM to improve through challenges produced by each other. We first define the standard reinforcement learning objective (Section~\ref{sec:prelim}), then describe how to select challenging inputs and update the branch that solves them (Section~\ref{sec:round}). Finally, we explain how selected outputs form the next epoch's input pools, allowing the model to generate new challenges as training progresses (Section~\ref{sec:evolve}).
\begin{figure}[t]
\centering
\includegraphics[width=\linewidth]{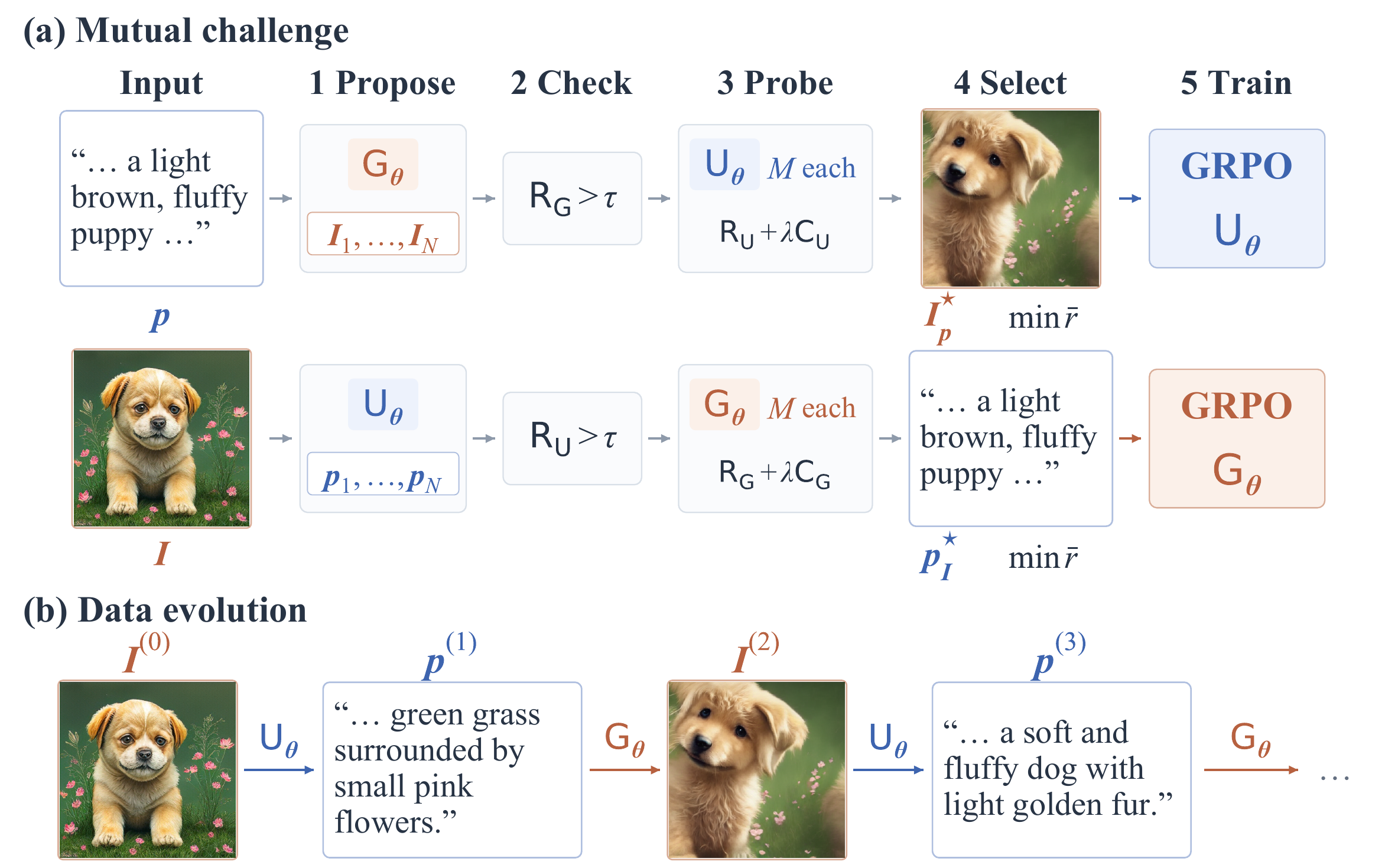}
\caption{\textbf{MATE: mutual challenge and data evolution.}
\textbf{(a) Mutual challenge.} The lower path exchanges the challenger and
solver roles of the upper path.
(1) The challenger proposes $N$ candidates. (2) A fixed scoring function
admits candidates above $\tau$, or the highest-scoring
candidate if none passes. (3) The solver makes $M$ attempts per admitted candidate,
scored by its scoring function $\mathsf{R}$ and input-consistency term
$\lambda\mathsf{C}$. (4) The lowest mean score $\bar r$ identifies the selected challenge.
(5) GRPO reuses all $M$ existing attempts for that challenge.
Inputs and selected challenges are records from (b), and other candidates
and attempts are shown symbolically.
\textbf{(b) Data evolution.} Consecutive image and text records from one
training trajectory illustrate $\mI^{(0)}\!\to\!\vp^{(1)}\!\to\!
\mI^{(2)}\!\to\!\vp^{(3)}$. Each arrow applies the procedure in (a) and
advances one epoch, so a selected output becomes the input of the opposite
branch. Quoted text is taken from actual model outputs and shortened for display.}
\label{fig:overview}
\end{figure}

\subsection{Preliminaries}\label{sec:prelim}
A UMM has generation and understanding branches that share a backbone. The generator $\mathsf{G}_{\vtheta}$ produces an image from a text prompt, and the understanding branch $\mathsf{U}_{\vtheta}$ produces a description of an image. Throughout, $\vp$ denotes a text prompt or description and $\mI$ denotes an image. In this paper, we consider post-training both branches with reinforcement learning, which updates a branch on scored samples of its own outputs. Outputs are scored by two fixed scoring functions, one per branch. The scoring function $\mathsf{R}_{\mathsf{G}}(\mI,\vp)$ measures the agreement of an image with a text, and $\mathsf{R}_{\mathsf{U}}(\vp,\mI)$ that of a text with an image. In both, the sampled output is written first and the reference second.

For a branch $\pi\in\{\mathsf{G},\mathsf{U}\}$ with an input $\vx$ and a reward function $\mathsf{r}$ on its outputs, the objective is the expected reward $\mathsf{J}(\vtheta;\pi,\vx,\mathsf{r})=\mathbb{E}_{\vy\sim\pi_{\vtheta}(\cdot\mid \vx)}[\mathsf{r}(\vy)]$, which is maximized over $\vtheta$ using sampled outputs. The standard procedure is direct post-training on a corpus $\mathcal{D}=\{(\vp_i,\mI_i)\}$. For every pair it maximizes $\mathsf{J}(\vtheta;\mathsf{G},\vp_i,\mathsf{R}_{\mathsf{G}}(\cdot,\vp_i))$ for the generator and $\mathsf{J}(\vtheta;\mathsf{U},\mI_i,\mathsf{R}_{\mathsf{U}}(\cdot,\mI_i))$ for the understanding branch, so each output is scored against the input it was produced from. The corpus fixes the inputs in advance, regardless of how well either branch already performs on them. MATE instead trains each branch on inputs that challenge it. It keeps the objective $\mathsf{J}$ and the optimizer of the standard procedure and differs from it only in where the inputs come from and what the outputs are scored against.

\subsection{Mutual challenge}\label{sec:round}
Direct post-training draws the inputs of a branch from the corpus. MATE instead seeks an input that challenges the branch under training. Since the two branches of a UMM are inverse tasks and the outputs of one are inputs of the other, the key idea of MATE is to treat those outputs as candidates and to train a branch on the candidate that yields its lowest score. One branch serves as the challenger that proposes candidates and the other as the solver that is trained on them (Figure~\ref{fig:overview}(a)). The roles can be exchanged, so the challenge is mutual.

\paragraph{Formulation.}
Take the understanding branch $\mathsf{U}_{\vtheta}$ as the challenger and the generator $\mathsf{G}_{\vtheta}$ as the solver. Given a source image $\mI$, the challenge is the candidate on which the solver has the lowest expected score,
\begin{equation}
\vp^{\star}_{\mI} \;=\; \argmin_{\vp \in \mathcal{V}_{\vtheta}(\mI)}\; \mathsf{s}_{\vtheta}(\vp; \mI),
\label{eq:ideal}
\end{equation}
where $\mathsf{s}_{\vtheta}(\vp;\mI)$ is the solver's expected reward on a candidate, and $\mathcal{V}_{\vtheta}(\mI)$ is the set of candidates the challenger may pose, consisting of its own descriptions that satisfy a consistency threshold with the source image,
\begin{equation}
\mathcal{V}_{\vtheta}(\mI) \;=\; \bigl\{\, \vp \in \operatorname{\mathsf{supp}}\, \mathsf{U}_{\vtheta}(\cdot\mid \mI) \;:\; \mathsf{R}_{\mathsf{U}}(\vp, \mI) > \tau \,\bigr\}.
\label{eq:moves}
\end{equation}
Candidates are outputs of the current challenger, and challenge selection operates on these outputs without training a separate adversary. We treat a candidate as valid if its agreement with the source image exceeds a threshold $\tau$. Applying this validity criterion before evaluating difficulty reduces the risk of selecting a candidate merely because it misrepresents the source image.

The solver's score is the expected reward of its attempts at the candidate, where each attempt is also rewarded for consistency with the source image,
\begin{equation}
\mathsf{s}_{\vtheta}(\vp; \mI) \;=\; \mathbb{E}_{\hat{\mI}\sim \mathsf{G}_{\vtheta}(\cdot\mid \vp)}\bigl[\, \mathsf{R}_{\mathsf{G}}(\hat{\mI},\vp) + \lambda\, \mathsf{C}_{\mathsf{G}}(\hat{\mI},\mI) \,\bigr],
\label{eq:score}
\end{equation}
where $\hat{\mI}$ is the solver's attempt at $\vp$, $\mathsf{C}_{\mathsf{G}}$ is a frozen consistency measure between two images, and $\lambda$ is its weight. The first term measures whether the generated image matches the candidate description, and the consistency term encourages it to preserve the content of the source image. This additional reference is available from the challenge construction and requires no new annotation.

When the understanding branch is the solver, the challenge starts from a source text $\vp$. The generator's candidates are its images $\mI$ with $\mathsf{R}_{\mathsf{G}}(\mI,\vp)>\tau$, and the understanding branch's score on a candidate is the expected value of $\mathsf{R}_{\mathsf{U}}(\hat{\vp}, \mI) + \lambda\, \mathsf{C}_{\mathsf{U}}(\hat{\vp}, \vp)$ over its descriptions $\hat{\vp}\sim \mathsf{U}_{\vtheta}(\cdot\mid \mI)$, where $\mathsf{C}_{\mathsf{U}}$ is the consistency measure between two texts.

\paragraph{Implementation.}
In practice, MATE does not solve Equation~\ref{eq:ideal} directly, because enumerating all possible challenger outputs is intractable. We approximate the search by sampling candidates and solver outputs. With the generator as the solver, a round starting from a source image $\mI$ samples $N$ candidates $\vp_1,\ldots,\vp_N \sim \mathsf{U}_{\vtheta}(\cdot\mid \mI)$ and admits $\mathcal{F}=\{\vp_n : \mathsf{R}_{\mathsf{U}}(\vp_n, \mI) > \tau\}$. If no candidate satisfies the threshold, $\mathcal{F}$ is set to contain only the candidate with the highest $\mathsf{R}_{\mathsf{U}}(\vp_n, \mI)$. For each admitted candidate $\vp$ it draws $M$ rollouts $\hat{\mI}_{m}\sim \mathsf{G}_{\vtheta}(\cdot\mid \vp)$ with rewards $r_{m}=\mathsf{R}_{\mathsf{G}}(\hat{\mI}_{m}, \vp) + \lambda\, \mathsf{C}_{\mathsf{G}}(\hat{\mI}_{m}, \mI)$, and their mean $\bar{r}(\vp) = \frac{1}{M}\sum_{m=1}^{M} r_{m}$ estimates the candidate's score of Equation~\ref{eq:score}. The round selects $\vp^{\star}_{\mI}=\argmin_{\vp \in \mathcal{F}} \bar{r}(\vp)$ as the challenge, approximating the search in Equation~\ref{eq:ideal} with sampled candidates and scores. The solver is then updated to maximize the objective $\mathsf{J}(\vtheta;\mathsf{G},\vp^{\star}_{\mI},\mathsf{r}^{\star})$ with the reward function $\mathsf{r}^{\star}=\mathsf{R}_{\mathsf{G}}(\cdot,\vp^{\star}_{\mI})+\lambda\,\mathsf{C}_{\mathsf{G}}(\cdot,\mI)$, so any policy-gradient method applies. We use GRPO~\citep{shao2024deepseekmath}, which treats the rollouts of an input as a group and standardizes their rewards within it. The $M$ rollouts already drawn for $\vp^{\star}_{\mI}$ serve as its group, so the same samples estimate the candidate's difficulty and provide the solver's training update. Appendix~\ref{app:algorithms} gives the details of the update and the full procedure.

\subsection{Self-play in data space}\label{sec:evolve}
Section~\ref{sec:round} defines a single round at a fixed source image or text. A fixed source bounds the difficulty that can be proposed from it, so rounds over a fixed corpus supply no further pressure once the model handles everything the corpus can pose. MATE therefore lets the inputs evolve with the model.

In MATE, self-play means that the model generates training challenges for its own two branches and uses the selected outputs to construct future challenges. Each branch runs one challenge-selection round for every input in its pool (Figure~\ref{fig:overview}(b)). Let $\mathcal{P}^{p}_{k}$ and $\mathcal{P}^{I}_{k}$ denote the pools of input texts and images at epoch $k$, initialized as $\mathcal{P}^{p}_{0}=\{\vp_i\}$ and $\mathcal{P}^{I}_{0}=\{\mI_i\}$ from the corpus. Each round selects a description $\vp^{\star}_{\mI}$ from a source image $\mI$, or an image $\mI^{\star}_{\vp}$ from a source text $\vp$. These selected candidates form the input pools for the next epoch,
\begin{equation}
\mathcal{P}^{I}_{k+1} \;=\; \bigl\{\, \mI^{\star}_{\vp} \;:\; \vp \in \mathcal{P}^{p}_{k} \,\bigr\},
\qquad
\mathcal{P}^{p}_{k+1} \;=\; \bigl\{\, \vp^{\star}_{\mI} \;:\; \mI \in \mathcal{P}^{I}_{k} \,\bigr\} ,
\label{eq:evolve}
\end{equation}
so an output selected for its difficulty to one branch becomes the source of new challenges for the other. Repeating this update produces a sequence that alternates between modalities, $\vp^{(0)} \to \mI^{(1)} \to \vp^{(2)} \to \cdots$, with one transition per epoch. Each round uses only a source text or image, so the procedure does not require paired data. Updating the pools allows the range of possible challenges to change with the model, whereas fixed inputs restrict the search to candidates derived from the original corpus.

We refer to this procedure as \emph{self-play in data space}: the evolving input pools and the current model's outputs supply the training challenges. Selected outputs carry information about each branch's current difficulties into the next epoch, allowing the challenge distribution to adapt without a separate opponent network or an archive of past checkpoints.

\section{Experiments}\label{sec:exp}

\subsection{Experimental setup}\label{sec:exp-setup}

\paragraph{Model \& data.}
We instantiate MATE on Janus-Pro-1B~\citep{chen2025januspro}, an autoregressive UMM with a 1.5B-parameter language backbone. We use 5000 image-text pairs from the text-to-image split of Align-Anything~\citep{ji2024alignanything} to initialize text and image pools. Training does not use the original pairings, and none of the corpus texts appears among the prompts of the generation benchmarks. Both scoring functions are implemented with the frozen reward model UnifiedReward~\citep{wang2025unified}, which takes no part in evaluation. The consistency measure is the similarity between SigLIP features. Appendix~\ref{app:exp} specifies the prompts and normalization of both scores.

\paragraph{Training.}
A round samples $N=4$ candidates and draws $M=8$ rollouts per admitted candidate, with threshold $\tau=0.5$, consistency weight $\lambda=0.1$, and standard-deviation floor $\epsilon=10^{-6}$. Both branches are updated by GRPO on all 2.1B model parameters with learning rate $10^{-6}$, clipping range $0.2$, and reference penalty $0.01$. Training runs for five epochs with an effective batch size of 48 on six GPUs. MATE takes about 29 hours because each input produces $N$ candidates and up to $NM$ rollouts. All training methods and ablations share the optimizer, scoring functions, corpus, and batch size. Each reported result comes from one training run, so we interpret sub-point GenEval differences cautiously.

\paragraph{Baseline methods.}
We compare against two groups of baselines. The first consists of the self-training methods UniRL~\citep{mao2025unirl} and SUDER~\citep{hong2025suder}, which we re-implement on Janus-Pro-1B using their published training procedures. The second consists of published generation, understanding, and unified models. We report their generation scores from their papers without prompt rewriting. For understanding, we use scores from the OpenVLM leaderboard, which uses the same evaluation toolkit, when available, and evaluate the remaining model--benchmark combinations ourselves. We evaluate Janus-Pro-1B and all three methods trained from it under the same protocol; bold marks the best result among these four.

\paragraph{Benchmarks.}
Generation is measured on GenEval~\citep{ghosh2023geneval} and DPG-Bench~\citep{hu2024ella}. We use the official GenEval evaluation code and the ELLA evaluation code for DPG-Bench, generating four images per prompt for each benchmark. GenEval reports the mean of its six category accuracies. For Janus-Pro-1B and its post-trained variants, images are sampled at 384 pixels with classifier-free guidance weight 5 and temperature 1. Understanding is measured with VLMEvalKit~\citep{duan2024vlmevalkit} using benchmark-specific scoring; Janus-based evaluations use the Janus-Pro chat template and greedy decoding. We report overall F1 on POPE~\citep{li2023pope}, the cognition score of MME~\citep{fu2023mme}, MMBench~\citep{liu2024mmbench}, the image split of SEED-Bench~\citep{li2024seedbench}, MMStar~\citep{chen2024mmstar}, V$^*$Bench~\citep{wu2024vstar}, the 2D and 3D splits of CV-Bench~\citep{tong2024cambrian}, and SEED-Bench-2-Plus~\citep{li2024seedbench2plus}. MMBench reports accuracy on the English development questions, which we evaluate for every model. The first five columns cover general understanding, and the last four assess fine-grained visual perception relevant to image description. The understanding average is the mean over the nine columns after the MME cognition score is rescaled to $[0,100]$ by its maximum of 800.

\subsection{Generation and understanding}\label{sec:exp-main}

\begin{table}[t]
\caption{Main results on image generation.}
\label{tab:main-gen}
\centering
\scriptsize
\setlength{\tabcolsep}{3pt}
\begin{tabular}{lccccccc@{\hspace{14pt}}cc}
\toprule
\multirow{2}{*}{Method} & \multirow{2}{*}{\# Params} & \multicolumn{6}{c@{\hspace{14pt}}}{GenEval$\uparrow$ (per category)} & \multicolumn{1}{c}{\textbf{GenEval$\uparrow$}} & DPG-Bench$\uparrow$ \\
\cmidrule(lr{13pt}){3-8}\cmidrule(lr){9-9}\cmidrule(lr){10-10}
 & & \makecell{Single\\Obj.} & \makecell{Two\\Obj.} & Counting & Colors & Position & \makecell{Color\\Attr.} & \textbf{Overall} & Overall \\
\midrule
\multicolumn{10}{l}{\textit{Generation only}} \\
SD3-Medium~\citep{esser2024sd3} & 2B & 99.0 & 94.0 & 72.0 & 89.0 & 33.0 & 60.0 & 74.0 & 84.1 \\
LlamaGen~\citep{sun2024llamagen} & 0.8B & 71.0 & 34.0 & 21.0 & 58.0 & 7.0 & 4.0 & 32.0 & -- \\
Sana-0.6B~\citep{xie2025sana} & 0.6B & 99.0 & 76.0 & 64.0 & 88.0 & 18.0 & 39.0 & 64.0 & 83.6 \\
Sana-1.6B~\citep{xie2025sana} & 1.6B & 99.0 & 77.0 & 62.0 & 88.0 & 21.0 & 47.0 & 66.0 & 84.8 \\
Lumina-Image 2.0~\citep{qin2025lumina2} & 2.6B & -- & 87.0 & 67.0 & -- & -- & 62.0 & 73.0 & 87.2 \\
\midrule
\multicolumn{10}{l}{\textit{UMMs}} \\
Janus~\citep{wu2025janus} & 1.3B & 97.0 & 68.0 & 30.0 & 84.0 & 46.0 & 42.0 & 61.0 & 79.7 \\
JanusFlow~\citep{ma2024janusflow} & 1.3B & 97.0 & 59.0 & 45.0 & 83.0 & 53.0 & 42.0 & 63.0 & 80.1 \\
Show-o~\citep{xie2024showo} & 1.3B & 98.0 & 80.0 & 66.0 & 84.0 & 31.0 & 50.0 & 68.0 & 67.3 \\
\addlinespace[3pt]
Janus-Pro-1B~\citep{chen2025januspro} & 1.5B & 98.4 & 84.1 & 53.1 & 91.2 & 63.8 & 56.0 & 74.4 & 82.1 \\
\quad + UniRL~\citep{mao2025unirl} & 1.5B & 97.5 & 82.8 & 51.6 & \textbf{92.0} & \textbf{66.5} & 60.5 & 75.2 & 82.1 \\
\quad + SUDER~\citep{hong2025suder} & 1.5B & \textbf{99.4} & 83.6 & 53.1 & 90.4 & 63.2 & 56.0 & 74.3 & 82.0 \\
\quad + MATE (ours) & 1.5B & 99.1 & \textbf{87.9} & \textbf{53.8} & 90.7 & 65.5 & \textbf{64.0} & \textbf{76.8} & \textbf{83.8} \\
\bottomrule
\end{tabular}
\end{table}

\begin{figure}[t]
\centering
\includegraphics[width=\linewidth]{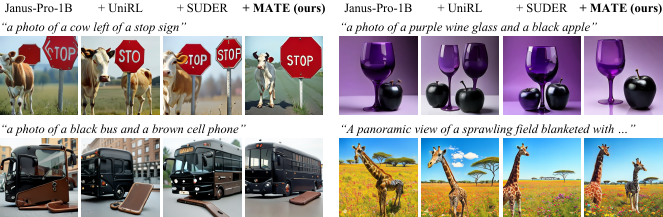}
\caption{\textbf{Generation samples.} Images for the same prompt from Janus-Pro-1B and the three self-training methods applied to it; three prompts are from GenEval and one from DPG-Bench. One selected sample per model and prompt is shown from four generated samples, using the same sampling settings. The DPG-Bench prompt asks for a giraffe and a zebra standing side by side in a field of wildflowers; its displayed text is shortened.}
\label{fig:exp-generation}
\end{figure}

\paragraph{Image generation.}
Table~\ref{tab:main-gen} reports generation results. MATE raises the GenEval score of Janus-Pro-1B by 2.4 points and its DPG-Bench score by 1.7, outperforming both self-training baselines on both benchmarks. The largest GenEval category gains are in two-object generation and color attribution, which require distinguishing objects and their attributes. Figure~\ref{fig:exp-generation} compares selected outputs for the same prompts across all four methods.

\begin{figure}[!htb]
\centering
\includegraphics[width=\linewidth]{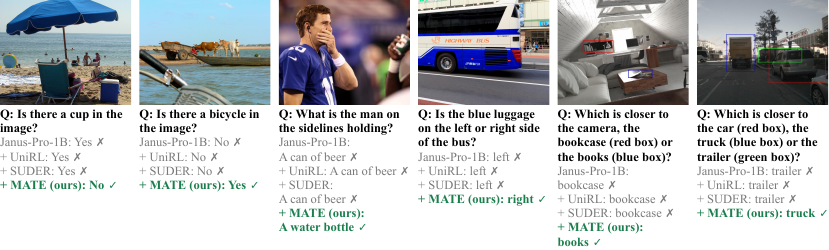}
\caption{\textbf{Understanding samples.} Six questions from POPE, SEED-Bench, V$^*$Bench, and CV-Bench. Multiple-choice answers show the selected option's text; three questions are shortened and images cropped for display. A check mark denotes agreement with the reference answer.}
\label{fig:exp-understanding}
\vspace{-10pt}
\end{figure}

\paragraph{Visual understanding.}
Table~\ref{tab:main-und} reports understanding results. MATE raises the average from 56.9 to 57.6 and improves eight of the nine benchmarks relative to the base model, with the largest gain on MME cognition; POPE decreases by 0.1 points. UniRL and SUDER fall below the base model on most understanding benchmarks. Together, these results show improvements in both image generation and average visual understanding under MATE. Figure~\ref{fig:exp-understanding} shows examples.

\begin{table}[!htbp]
\caption{Main results on visual understanding. Higher is better on every benchmark.}
\label{tab:main-und}
\centering
\scriptsize
\setlength{\tabcolsep}{2.5pt}
\begin{tabular}{lcccccccccc@{\hspace{6pt}}c}
\toprule
Method & \# Params & POPE & MME-C & MMB & SEED & MMStar & V$^*$Bench & CV-2D & CV-3D & SEED2+ & \textbf{Avg.} \\
\cmidrule(r{4pt}){1-11}\cmidrule(l){12-12}
\multicolumn{12}{l}{\textit{Understanding only}} \\
LLaVA-v1.5~\citep{liu2024llava15} & 7B & 86.1 & 302.1 & 72.9 & 65.8 & 33.1 & 45.0 & 62.0 & 64.3 & 41.3 & 56.5 \\
DeepSeek-VL-1.3B~\citep{lu2024deepseekvl} & 1.3B & 85.9 & 225.0 & 74.5 & 66.0 & 39.9 & 42.9 & 66.2 & 62.7 & 43.7 & 56.7 \\
LLaVA-OV-0.5B~\citep{li2024llavaonevision} & 0.5B & 87.8 & 237.5 & 67.1 & 66.6 & 37.7 & 39.3 & 38.6 & 50.6 & 52.6 & 52.2 \\
\midrule
\multicolumn{12}{l}{\textit{UMMs}} \\
Janus~\citep{wu2025janus} & 1.3B & 84.1 & 238.6 & 70.0 & 62.9 & 37.6 & 41.4 & 49.1 & 58.7 & 38.5 & 52.5 \\
JanusFlow~\citep{ma2024janusflow} & 1.3B & 85.4 & 268.2 & 76.9 & 70.3 & 47.2 & 44.5 & 61.9 & 67.8 & 45.9 & 59.3 \\
Show-o~\citep{xie2024showo} & 1.3B & 82.4 & 238.2 & 50.9 & 54.6 & 29.6 & 37.7 & 34.2 & 49.1 & 23.3 & 43.5 \\
\addlinespace[3pt]
Janus-Pro-1B~\citep{chen2025januspro} & 1.5B & \textbf{84.3} & 263.6 & 74.5 & 66.9 & 43.5 & 40.8 & 59.8 & 58.9 & 50.9 & 56.9 \\
\quad + UniRL~\citep{mao2025unirl} & 1.5B & 83.8 & 258.9 & 73.4 & 65.7 & 42.0 & 42.4 & 57.0 & 55.1 & 49.2 & 55.7 \\
\quad + SUDER~\citep{hong2025suder} & 1.5B & 83.5 & 255.4 & 73.5 & 65.8 & \textbf{44.5} & 40.3 & 59.6 & 56.8 & 50.5 & 56.3 \\
\quad + MATE (ours) & 1.5B & 84.2 & \textbf{281.4} & \textbf{74.7} & \textbf{67.0} & 43.7 & \textbf{42.9} & \textbf{60.5} & \textbf{59.2} & \textbf{51.2} & \textbf{57.6} \\
\bottomrule
\end{tabular}
\end{table}

\subsection{Ablation and analysis}\label{sec:exp-ablation}

\paragraph{Challenge selection and evolving inputs.}
Figure~\ref{fig:cycle}(b) compares MATE with three variants. It reports the average over the 9 understanding benchmarks (Table~\ref{tab:main-und}) and the average of GenEval and DPG-Bench. Keeping the initial pools fixed lowers the generation average of MATE from 80.3 to 79.9, and additionally selecting the highest-scoring candidate instead of the hardest (cooperative selection) lowers it to 79.5, compared with 78.3 for the base model. The understanding averages of MATE and these two variants range from 57.6 to 57.8, all above the base model's 56.9. Thus, challenge selection and evolving inputs account for part of the generation gain, whereas the understanding gain does not depend on them. Each difference is 0.4 points and comes from a single run, so these results are suggestive rather than conclusive. Removing the admission threshold from MATE, so that candidates are not checked against $\tau$ as in Equation~\ref{eq:moves}, lowers both averages, to 57.3 and 79.3. Table~\ref{tab:ablation-full} in Appendix~\ref{app:exp} evaluates three more variants, which remove the consistency term, train on the groups of all admitted candidates, or use separate networks for the two branches.

\paragraph{Evolving training examples.}
Figure~\ref{fig:evolution} follows one example through five epochs, and Figure~\ref{fig:evolution-more} in Appendix~\ref{app:exp} shows a second. The selected images and descriptions change as the input pools are updated, illustrating how MATE constructs subsequent inputs from its own outputs.

\begin{figure}[t]
\centering
\includegraphics[width=.85\linewidth]{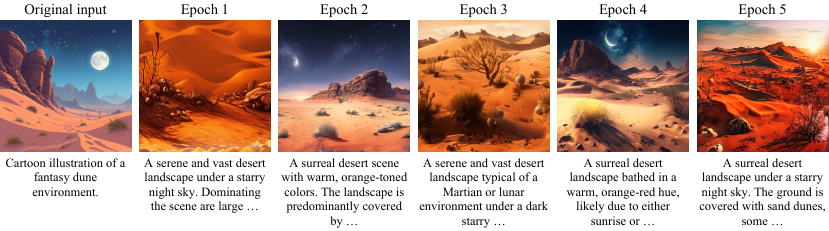}
\caption{\textbf{An evolving training input under MATE.} An original dataset pair (left) is followed through five epochs. Selected images and descriptions become inputs for the next epoch. The image and text in each column come from separate training directions and need not form a matched pair. Descriptions are shortened for display. Figure~\ref{fig:evolution-more} in Appendix~\ref{app:exp} shows a second example.}
\label{fig:evolution}
\end{figure}

\newsavebox{\cycleplotbox}
\begin{figure}[!htb]
\sbox{\cycleplotbox}{\includegraphics[width=0.51\linewidth]{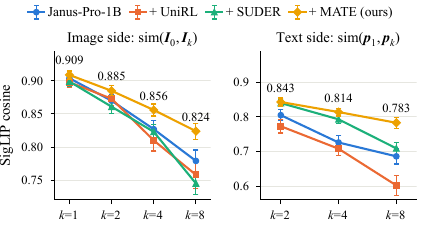}}%
\begin{minipage}[t]{0.51\linewidth}
\vspace{0pt}
\centering
{\small\textbf{(a) Round-trip consistency}}\par\smallskip
\usebox{\cycleplotbox}
\end{minipage}\hfill
\begin{minipage}[t]{0.46\linewidth}
\vspace{0pt}
\centering
{\small\textbf{(b) Ablation of MATE}}\par\smallskip
\begin{minipage}[t][\ht\cycleplotbox][c]{\linewidth}
\centering
\scriptsize
\setlength{\tabcolsep}{3pt}
\begin{tabular}{lcc}
\toprule
 & Und.\ Avg.$\uparrow$ & Gen.\ Avg.$\uparrow$ \\
\midrule
Janus-Pro-1B & 56.9 & 78.3 \\
MATE (ours) & 57.6 & 80.3 \\
\midrule
MATE w/o evolving pools & 57.7 & 79.9 \\
MATE w/o adversarial training & 57.8 & 79.5 \\
MATE w/o admission threshold & 57.3 & 79.3 \\
\bottomrule
\end{tabular}

\end{minipage}
\end{minipage}
\vspace{-10pt}
\caption{\textbf{Round-trip consistency and ablation.} (a) Left: SigLIP cosine between the original image $\mI_0$ and the image $\mI_k$ after $k$ rounds. Right: SigLIP text cosine between the first description $\vp_1$ and the $k$-th description $\vp_k$. Points are means over 32 held-out samples and whiskers are standard errors. (b) Understanding and generation averages of Janus-Pro-1B, MATE, and three variants of MATE.}
\label{fig:cycle}
\vspace{-10pt}
\end{figure}

\paragraph{Round-trip consistency.}\label{sec:exp-cycle}
Starting from 32 held-out Align-Anything test images, we apply $k$ rounds of description and regeneration. Figure~\ref{fig:cycle}(a) reports two mean cosine similarities, each computed with one fixed encoder for all methods. On the image side, we compare the original image with the image after $k$ rounds, using the SigLIP~\citep{zhai2023siglip} vision encoder of the base UMM. On the text side, we compare the first description with the $k$-th description, using the text encoder of the same SigLIP model family. Appendix~\ref{app:exp} gives further details. On the image side, MATE exceeds the base model at every evaluated number of rounds. After eight rounds, its similarity is 0.824, compared with 0.779 for the base model, 0.758 for UniRL, and 0.745 for SUDER. On the text side, MATE also exceeds all three at every evaluated number of rounds, with 0.783 after eight rounds compared with 0.686 for the base model, 0.602 for UniRL, and 0.709 for SUDER. These results indicate that MATE better preserves content over repeated image--text cycles on both sides. Because MATE's consistency term is also a SigLIP similarity, this evaluation is not fully independent of its training reward.

\section{Conclusion}
We present MATE, a post-training framework that uses the generation and understanding branches of a UMM to challenge each other. By checking candidates for source consistency and selecting those the opposite branch handles worst, MATE focuses training on the model's current weaknesses without a separately trained adversary or new annotations. Selected images and descriptions become the next epoch's inputs, allowing challenges to evolve with the model. On Janus-Pro-1B, MATE improves both image generation and average visual understanding over the base model and two self-training baselines, while better preserving content through repeated image--text cycles. Ablations indicate that hard challenge selection and evolving inputs contribute to the generation gains.

\paragraph{Limitations.} Due to computational constraints, our experiments use a 1B-scale UMM and 5000 training pairs, and we plan to evaluate MATE with larger UMMs and larger, higher-quality training datasets in future work. MATE also approximates the search in Equation~\ref{eq:ideal} with a few sampled candidates and rollouts, so the selected challenge may not be the optimal one. Proposing and scoring these candidates also adds computation relative to other post-training methods, which more efficient candidate screening could reduce.

\subsection*{AI use statement}

In this work, we used generative AI tools (LLM-based coding assistants) to
implement and debug parts of the training and evaluation code, to write
scripts for launching, monitoring, and aggregating experiments, to draft data
analysis scripts, and to edit the manuscript for readability. We have not used
generative AI tools to formulate the research hypotheses, to design the
method, or to generate synthetic training data beyond the self-play procedure
described in Section~\ref{sec:method}; assisting in the writing of proofs is
not applicable to this work. We have reviewed all AI-assisted work:
LLM-generated code was inspected and tested by the authors, all reported
numbers were recomputed from raw evaluation outputs under the protocol in
Section~\ref{sec:exp}, and all text was checked by the authors. We take
responsibility for the final content of this work, including text, claims, and
artifacts produced with the aid of generative AI.

\subsection*{Reproducibility statement}

All experiments build on publicly available models and datasets. The training
recipe, hyperparameters, and the evaluation protocol (official GenEval and
DPG-Bench evaluation code, VLMEvalKit with the Janus-Pro chat template and
greedy decoding) are described in Section~\ref{sec:exp} and the appendix; training
scripts, configuration files, and evaluation code will be released as
supplementary material, together with the recorded candidate scores and
selections of the self-play loop used for the qualitative analysis.

\bibliography{references}
\bibliographystyle{iclr2027_conference}

\clearpage
\appendix
\raggedbottom
\section*{Appendix}
This appendix supplements the main text in three parts. Appendix~\ref{app:positioning} compares MATE with the closest post-training methods for UMMs. Appendix~\ref{app:algorithms} gives the full training procedure, the two directions of a round, and the GRPO update. Appendix~\ref{app:exp} specifies the scoring functions and hyperparameters and reports additional variants of MATE, a second example of evolving training inputs, and details of the round-trip evaluation.

\section{Positioning against prior work}\label{app:positioning}
Table~\ref{tab:positioning} compares MATE with the post-training methods for UMMs discussed in Section~\ref{sec:related}, which are the cooperative self-training methods and UniGame. The column headed \emph{Other branch supplies} states what one branch contributes to the training of the other. In the cooperative methods, this branch scores the outputs of the other through the likelihood that it assigns to the original input, the correctness of its answers to questions derived from the prompt, or a rating that it writes. In UniGame and MATE it supplies challenges instead, and the scores in MATE come from fixed scoring functions (Appendix~\ref{app:exp}). \emph{Tasks optimized} lists the tasks with training objectives, where $\mathsf{G}$ and $\mathsf{U}$ denote generation and understanding; shared parameters can affect both capabilities. \emph{Training inputs} states where the training inputs come from. A corpus is fixed before training, constructed prompts are built so that the answers to their questions are known, and model-written prompts are produced by the model under training. UniGame adds a buffer that replays hard examples, whereas MATE rebuilds its input pools every epoch from the outputs selected in the previous one. \emph{Hard cases selected for} lists the tasks whose training inputs are selected by solver difficulty, and \emph{Added module} names any trainable module that a method adds to the model.
\begin{table}[htb]
\centering
\caption{Comparison of MATE with the closest post-training methods for UMMs. Hard-case selection refers to choosing training inputs by the difficulty they pose to the task being optimized.}
\label{tab:positioning}
\scriptsize\setlength{\tabcolsep}{5pt}
\begin{tabular}{@{}llclcc@{}}
\toprule
Method & \makecell[l]{Other branch\\supplies} & \makecell{Tasks\\optimized} & \makecell[l]{Training\\inputs} & \makecell{Hard cases\\selected for} & \makecell{Added\\module} \\
\midrule
SUDER~\citep{hong2025suder}           & likelihood & $\mathsf{G}$, $\mathsf{U}$ & corpus                    & --   & -- \\
GvU~\citep{gvu2026}                   & likelihood & $\mathsf{G}$ & corpus                    & --   & -- \\
UniRL~\citep{mao2025unirl}            & answers    & $\mathsf{G}$, $\mathsf{U}$ & constructed prompts    & --   & -- \\
SILMM~\citep{qu2024silmm}             & answers    & $\mathsf{G}$ & model-written prompts     & --   & -- \\
HermesFlow~\citep{yang2025hermesflow} & answers    & $\mathsf{G}$, $\mathsf{U}$ & corpus                    & --   & -- \\
SRUM~\citep{jin2025srum}              & ratings    & $\mathsf{G}$ & corpus                    & --   & -- \\
UniCorn~\citep{han2026unicorn}        & ratings    & $\mathsf{G}$, $\mathsf{U}$ & model-written prompts     & --   & -- \\
\midrule
UniGame~\citep{su2025unigame}         & challenges & $\mathsf{U}$ & corpus, replay buffer & $\mathsf{U}$ & perturber \\
\midrule
\textbf{MATE (ours)}                  & challenges & $\mathsf{G}$, $\mathsf{U}$ & evolving pools            & $\mathsf{G}$, $\mathsf{U}$ & -- \\
\bottomrule
\end{tabular}
\end{table}

\section{Additional details on the algorithm}\label{app:algorithms}
Algorithm~\ref{alg:selfplay-main} is the full procedure of Section~\ref{sec:method}. Algorithms~\ref{alg:selfplay-understand} and~\ref{alg:selfplay-generate} give the round of Section~\ref{sec:round} in its two directions. In \textsc{ForwardLoop} the round starts from a source text, the generator is the challenger, and the understanding branch is the solver. \textsc{ReverseLoop} starts from a source image and exchanges the two roles, which is the case formulated in Equations~\ref{eq:ideal} to~\ref{eq:score}. Each procedure returns the selected candidate together with its group of $M$ rollouts, and Algorithm~\ref{alg:selfplay-main} updates the solver with GRPO on the returned groups.
\begin{ruledalgorithm}[alg:selfplay-main]{MATE: one training run}
\Require input pools $\mathcal{P}^{p}_{0}$ and $\mathcal{P}^{I}_{0}$ drawn from $\mathcal{D}$, UMM $\vtheta$, epoch count $E$, floor $\epsilon$
\For{$k=0$ to $E-1$} \Comment{update both branches in each epoch}
    \State For each $\vp\in\mathcal{P}^{p}_{k}$: \; $(\mI^{\star}_{\vp},\, g_{\vp}) \gets \textsc{ForwardLoop}(\vp,\vtheta)$ \Comment{hard images for $\mathsf{U}_{\vtheta}$}
    \State For each $\mI\in\mathcal{P}^{I}_{k}$: \; $(\vp^{\star}_{\mI},\, g_{\mI}) \gets \textsc{ReverseLoop}(\mI,\vtheta)$ \Comment{hard descriptions for $\mathsf{G}_{\vtheta}$}
    \State Update $\vtheta$ with GRPO on the groups $\{g_{\vp}\}$ whose reward standard deviation is at least $\epsilon$, then likewise on $\{g_{\mI}\}$
    \State $\mathcal{P}^{I}_{k+1} \gets \{\mI^{\star}_{\vp} : \vp\in\mathcal{P}^{p}_{k}\}$, \quad $\mathcal{P}^{p}_{k+1} \gets \{\vp^{\star}_{\mI} : \mI\in\mathcal{P}^{I}_{k}\}$ \Comment{inputs of the other loop}
\EndFor
\State \Return updated model $\vtheta$
\end{ruledalgorithm}

Let $r_1,\ldots,r_M$ denote the rollout rewards of a returned group, $\bar{r}$ their mean, and $\sigma$ their standard deviation. The group-relative advantage of rollout $m$ is
\begin{equation}
A_m=\frac{r_{m}-\bar{r}}{\sigma}.
\label{eq:grpo}
\end{equation}
The update then follows the clipped GRPO objective of \citet{shao2024deepseekmath} with these advantages and a KL penalty toward a reference policy, and Table~\ref{tab:hparams} lists the clipping range and the penalty weight. A group whose standard deviation falls below the floor $\epsilon$ is excluded from the update, since its rewards are nearly identical and their standardized advantages carry no signal. DAPO~\citep{yu2025dapo} likewise filters groups with no reward variation, whereas other work reshapes their advantages~\citep{rlzvp2025} or returns their queries to the same policy~\citep{queryrecycling2026}. In MATE the selected candidate of an excluded group still enters the next epoch's input pool as a source for the other branch, so the floor does not change challenge selection. Rollouts from unselected candidates are used only to estimate their difficulty. Training on the groups of all admitted candidates instead is evaluated in Table~\ref{tab:ablation-full}.
\begin{ruledalgorithm}[alg:selfplay-understand]{\textsc{ForwardLoop}: selecting challenging images for the understanding branch}
\Require source text $\vp$, UMM $\vtheta$, scoring functions $\mathsf{R}_{\mathsf{G}}$ and $\mathsf{R}_{\mathsf{U}}$, consistency measure $\mathsf{C}_{\mathsf{U}}$, weight $\lambda$, threshold $\tau$, candidate count $N$, rollout count $M$
\For{$n=1$ to $N$}
    \State Sample candidate image $\mI_n \sim \mathsf{G}_{\vtheta}(\cdot\mid \vp)$
    \State Compute admission score $\mathsf{R}_{\mathsf{G}}(\mI_n, \vp)$
\EndFor
\State Form admitted set $\mathcal{F}=\{n : \mathsf{R}_{\mathsf{G}}(\mI_n,\vp) > \tau\}$
\If{$\mathcal{F}=\emptyset$}
    \State $\mathcal{F}\gets\{\argmax_n \mathsf{R}_{\mathsf{G}}(\mI_n,\vp)\}$
\EndIf
\For{each $n\in\mathcal{F}$}
    \For{$m=1$ to $M$}
        \State Sample description $\hat{\vp}_{n,m} \sim \mathsf{U}_{\vtheta}(\cdot\mid \mI_n)$
        \State $r_{n,m} \gets \mathsf{R}_{\mathsf{U}}(\hat{\vp}_{n,m}, \mI_n) + \lambda\, \mathsf{C}_{\mathsf{U}}(\hat{\vp}_{n,m}, \vp)$
    \EndFor
    \State $\bar{r}_n \gets \frac{1}{M}\sum_{m=1}^{M} r_{n,m}$ \Comment{mean solver score on candidate $\mI_n$}
\EndFor
\State $n^{\star} \gets \argmin_{n\in\mathcal{F}} \bar{r}_n$
\State \Return selected candidate $\mI_{n^{\star}}$ and its group $\{(\hat{\vp}_{n^{\star},m}, r_{n^{\star},m})\}_{m=1}^{M}$
\end{ruledalgorithm}

\begin{ruledalgorithm}[alg:selfplay-generate]{\textsc{ReverseLoop}: selecting challenging descriptions for the generator}
\Require source image $\mI$, UMM $\vtheta$, scoring functions $\mathsf{R}_{\mathsf{G}}$ and $\mathsf{R}_{\mathsf{U}}$, consistency measure $\mathsf{C}_{\mathsf{G}}$, weight $\lambda$, threshold $\tau$, candidate count $N$, rollout count $M$
\For{$n=1$ to $N$}
    \State Sample candidate description $\vp_n \sim \mathsf{U}_{\vtheta}(\cdot\mid \mI)$
    \State Compute admission score $\mathsf{R}_{\mathsf{U}}(\vp_n, \mI)$
\EndFor
\State Form admitted set $\mathcal{F}=\{n : \mathsf{R}_{\mathsf{U}}(\vp_n,\mI) > \tau\}$
\If{$\mathcal{F}=\emptyset$}
    \State $\mathcal{F}\gets\{\argmax_n \mathsf{R}_{\mathsf{U}}(\vp_n,\mI)\}$
\EndIf
\For{each $n\in\mathcal{F}$}
    \For{$m=1$ to $M$}
        \State Sample image $\hat{\mI}_{n,m} \sim \mathsf{G}_{\vtheta}(\cdot\mid \vp_n)$
        \State $r_{n,m} \gets \mathsf{R}_{\mathsf{G}}(\hat{\mI}_{n,m}, \vp_n) + \lambda\, \mathsf{C}_{\mathsf{G}}(\hat{\mI}_{n,m}, \mI)$
    \EndFor
    \State $\bar{r}_n \gets \frac{1}{M}\sum_{m=1}^{M} r_{n,m}$ \Comment{mean solver score on candidate $\vp_n$}
\EndFor
\State $n^{\star} \gets \argmin_{n\in\mathcal{F}} \bar{r}_n$
\State \Return selected candidate $\vp_{n^{\star}}$ and its group $\{(\hat{\mI}_{n^{\star},m}, r_{n^{\star},m})\}_{m=1}^{M}$
\end{ruledalgorithm}

\section{Experimental details}\label{app:exp}
\paragraph{Scoring functions.}
UnifiedReward~\citep{wang2025unified} implements both scoring functions (Section~\ref{sec:exp-setup}); we use the released checkpoint \texttt{UnifiedReward-qwen-7b}, built on Qwen2.5-VL-7B-Instruct, with greedy decoding. Its pointwise image-against-text scores define $\mathsf{R}_{\mathsf{G}}$ on a 1--5 scale, and its pointwise text-against-image scores define $\mathsf{R}_{\mathsf{U}}$ on a 0--100 scale. We use the released pointwise prompts unchanged across all runs and rescale both scores to $[0,1]$ before they are used. The same scores decide admission against $\tau$. UnifiedReward takes no part in evaluation.

\paragraph{Hyperparameters and parameter counts.}
Table~\ref{tab:hparams} lists the training hyperparameters and the evaluation settings of Section~\ref{sec:exp}. The parameter counts in Tables~\ref{tab:main-gen} and~\ref{tab:main-und} are the language-model parameters for UMMs and understanding-only models and the total parameters for generation-only models, as reported in their respective papers.
\begin{table}[htb]
\caption{Training hyperparameters and evaluation settings.}
\label{tab:hparams}
\centering
\scriptsize
\setlength{\tabcolsep}{5pt}
\begin{tabular}{llll}
\toprule
\multicolumn{4}{l}{\textit{Training}} \\
Candidates per input $N$ & 4 & Learning rate & $10^{-6}$ \\
Rollouts per candidate $M$ & 8 & GRPO clipping range & 0.2 \\
Admission threshold $\tau$ & 0.5 & Reference penalty & 0.01 (low-variance KL) \\
Standard-deviation floor $\epsilon$ & $10^{-6}$ & Trainable parameters & all (2.1B) \\
Epochs & 5 & Hardware & 6$\times$ NVIDIA B200 \\
Effective batch size & 48 (24 inputs per step) & Consistency weight $\lambda$ & 0.1 \\
\midrule
\multicolumn{4}{l}{\textit{Evaluation}} \\
Image sampling & \multicolumn{3}{l}{384 px, guidance weight 5, temperature 1, 4 images per prompt} \\
Text decoding & \multicolumn{3}{l}{greedy} \\
\bottomrule
\end{tabular}
\end{table}

\paragraph{Additional variants.}
Table~\ref{tab:ablation-full} reports the understanding and generation averages of three additional variants of MATE, together with Janus-Pro-1B and MATE for reference. Both averages are defined in Section~\ref{sec:exp} and are computed before rounding. The first variant removes the consistency term, setting $\lambda=0$ in Equation~\ref{eq:score}. Without it, both averages remain within 0.2 points of those of MATE, so at this scale the term has little effect on them. The next variant trains on the groups of all admitted candidates instead of the selected group alone, which gives up to $N$ groups per source rather than one. Its averages are also within 0.3 points of those of MATE. The last variant runs MATE on two separate networks, using Qwen2-VL-2B~\citep{wang2024qwen2vl} for understanding and SimpleAR-0.5B~\citep{wang2025simplear} for generation. The pair has 2.8B parameters in total, compared with 2.1B for the complete Janus-Pro-1B model. Relative to its own base models, the pair gains 2.1 points on the generation average and loses 0.9 points on the understanding average. MATE improves both averages with a shared backbone, including a 0.7-point gain in understanding. This comparison is consistent with a benefit from shared parameters, although the systems also differ in their base models and total parameter counts.
\begin{table}[!htb]
\caption{Additional variants of MATE, with Janus-Pro-1B and MATE for reference.}
\label{tab:ablation-full}
\centering
\scriptsize
\setlength{\tabcolsep}{3pt}
\begin{tabular}{lcc}
\toprule
 & Und.\ Avg.$\uparrow$ & Gen.\ Avg.$\uparrow$ \\
\midrule
Janus-Pro-1B & 56.9 & 78.3 \\
MATE & 57.6 & 80.3 \\
\midrule
w/o consistency term & 57.5 & 80.5 \\
All admitted candidates & 57.5 & 80.6 \\
Two separate networks & 66.3 & 64.1 \\
\quad $\Delta$ vs.\ own base models & $-$0.9 & $+$2.1 \\
\bottomrule
\end{tabular}
\end{table}

\begin{figure}[!htb]
\centering
\includegraphics[width=\linewidth]{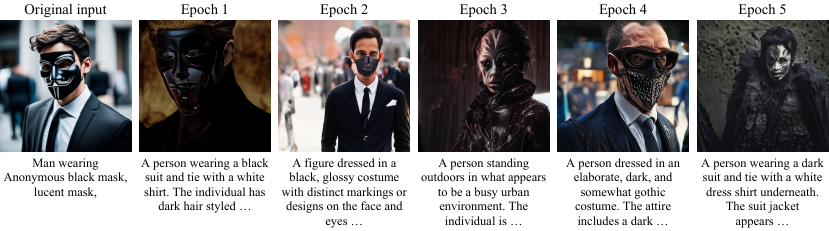}
\caption{\textbf{A second evolving training input under MATE,} displayed as in Figure~\ref{fig:evolution}. Descriptions are shortened for display.}
\label{fig:evolution-more}
\end{figure}

\paragraph{Evolving training examples.}
Figure~\ref{fig:evolution-more} follows a second original dataset pair through five epochs of MATE training, complementing the example in Figure~\ref{fig:evolution}.

\begin{figure}[!htb]
\centering
\includegraphics[width=\linewidth]{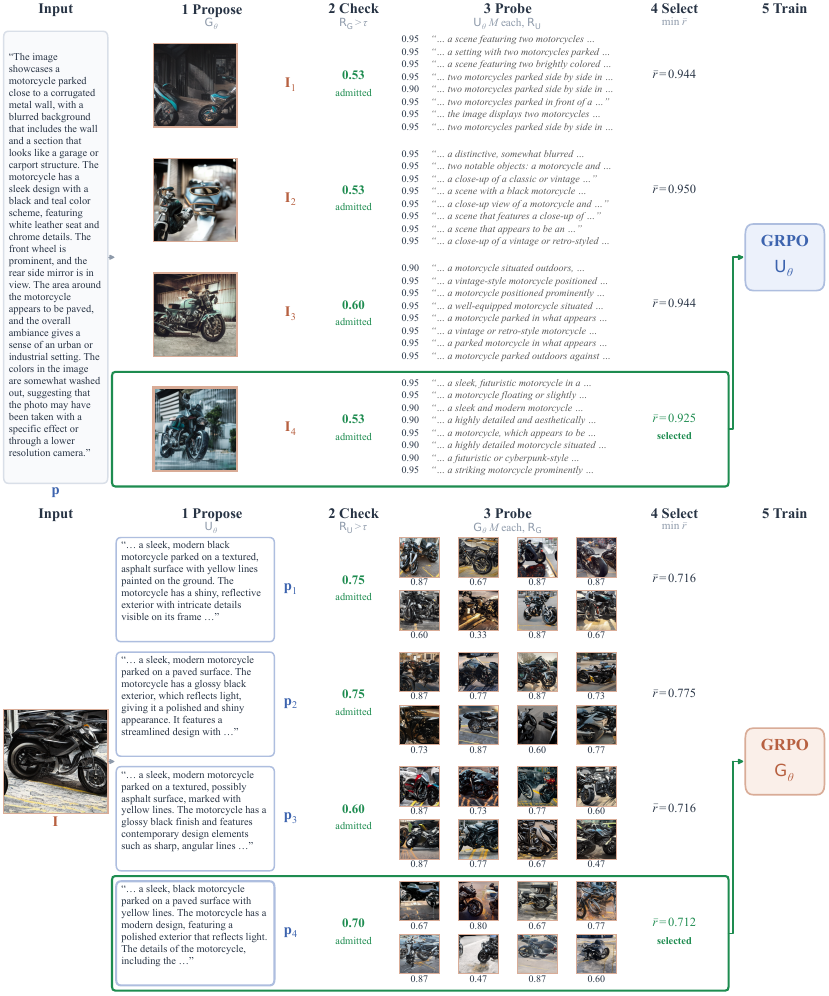}
\caption{\textbf{One complete round of MATE, in the format of Figure~\ref{fig:overview}(a).} All candidates, attempts, and scores of one training pair in one epoch (pair 4, epoch 2 of the main run). Upper row: from the source description $\vp$, the generation branch proposes $N=4$ images $\mI_1,\dots,\mI_4$; each is checked against $\vp$ ($\mathsf{R}_{\mathsf{G}}$, admission threshold $\tau=0.5$); the understanding branch writes $M=8$ descriptions of each admitted image, scored by $\mathsf{R}_{\mathsf{U}}$ (the eight scores and one example are shown); the image with the lowest mean score $\bar r$ is selected and its eight attempts train $\mathsf{U}_{\vtheta}$ with GRPO. Lower row: from the source image $\mI$, the understanding branch proposes $N=4$ descriptions $\vp_1,\dots,\vp_4$, checked against $\mI$ ($\mathsf{R}_{\mathsf{U}}$); the generation branch produces $M=8$ images from each admitted description, scored by $\mathsf{R}_{\mathsf{G}}$; the description with the lowest $\bar r$ is selected and its attempts train $\mathsf{G}_{\vtheta}$. The two selected outputs become this pair's inputs in the next epoch. Descriptions are shortened for display.}
\label{fig:round}
\end{figure}

\paragraph{A complete round.}
Figure~\ref{fig:round} shows every candidate, attempt, and score of one training pair in one epoch, laid out stage by stage as in Figure~\ref{fig:overview}(a) for both directions of the round; the two selected outputs become the pair's inputs in the next epoch.

\paragraph{Round-trip evaluation.}
This paragraph specifies the encoders and two details of the evaluation in Section~\ref{sec:exp-cycle}. The image-side similarity uses the SigLIP-L/16-384 vision encoder~\citep{zhai2023siglip} bundled with Janus-Pro-1B~\citep{chen2025januspro}: each image is resized to 384 pixels, the 576 patch embeddings are averaged, and the result is $\ell_2$-normalized. The text-side similarity uses the text encoder of the released \texttt{siglip-large-patch16-384} checkpoint~\citep{zhai2023siglip}. It compares the first description $\vp_1$ with the description $\vp_k$ of the image produced in round $k-1$, so it is defined from $k=2$. A description longer than the 64-token window of the SigLIP text encoder is split into windows, and the window embeddings are averaged. The 32 held-out samples are the first 32 pairs of the Align-Anything test split; descriptions are decoded greedily with at most 256 tokens, and images are sampled with the settings of Table~\ref{tab:hparams}.

\end{document}